\documentclass{article}

\PassOptionsToPackage{numbers,sort&compress}{natbib}
\usepackage[final]{ewrl_2026} 

\usepackage[utf8]{inputenc}
\usepackage[T1]{fontenc}
\usepackage{hyperref}
\hypersetup{hidelinks}
\usepackage{url}
\usepackage{booktabs}
\usepackage{amsfonts}
\usepackage{amsmath}
\usepackage{nicefrac}
\usepackage{microtype}
\usepackage{xcolor}
\usepackage{graphicx}
\usepackage{float}

\title{State Representation Matters in Deep Reinforcement Learning: Application to Energy Trading}

\author{
Jesper Klicks\thanks{Equal contribution. Email: \href{mailto:jesper.klicks@example.com}{j.j.klicks@student.vu.nl}},
Sander Vr\v{z}ina\footnotemark[1]\thanks{Equal contribution. Email: \href{mailto:sander.vrzina@example.com}{s.vrzina@student.vu.nl}},
Vincent Fran\c{c}ois-Lavet\thanks{Email: \href{mailto:vincent.françois-lavet@example.com}{vincent.francoislavet@vu.nl}}
}

\date{}

\begin{document}

\maketitle

\begin{abstract}
Energy trading decisions depend not only on current market prices, but also on expected future market conditions, and operational constraints. This makes the state representation given to a reinforcement learning agent an important design choice. We study this in HydroDam, a pumped-storage arbitrage environment, using a fixed Double DQN agent. The environment, action space, reward function, network, and training protocol are kept fixed; only the market features are changed. We compare absolute price/calendar features, relative features that compare current prices with recent market history, forecast features, and all combinations of these three feature families. Policies are trained and selected using 2007--2011 Belgian day-ahead prices and evaluated on two test settings: a later same-market test set from 2012--2025 and 39 other ENTSO-E market zones. Absolute features look strong on validation but fall to 28.8\% on the test set and a median 5.7\% across zones. Relative-only and forecast-only states also stay below a rolling price-score heuristic in the cross-zone median. Combining feature families is much stronger: absolute + relative reaches 49.9\% on the test set and a 39.8\% cross-zone median, while absolute + relative + forecast reaches 55.6\% and 47.5\%. These results suggest that state representation is not a minor preprocessing choice in storage-trading RL, but a central part of the policy design: robust transfer requires combining price scale, recent relative price context, and short-horizon forecast information, rather than relying on any single feature family.
\end{abstract}

\noindent\textbf{Source code:} available at \url{https://github.com/Fluxons/hydrodam}.

\section{Introduction}

Energy trading is a sequential decision problem under uncertainty. In the context of trading energy based on a dam for instance, each pumping or release decision changes the immediate revenue, but also changes the reservoir state available for later hours. Reinforcement learning is a natural fit for this setting because the agent can learn a trading policy by interacting with a simulated storage environment. However, the quality of the learned policy depends heavily on the features that are given to the agent. Price, calendar, historical-window, and forecast features all contain useful information, but they describe the market in different ways. Choosing which of these feature types to include can therefore strongly affect both the trading behavior and the generalization of the model.

This paper asks:
\begin{quote}
How do absolute, relative, and forecast market features affect the performance and generalization of model-free reinforcement learning for energy trading?
\end{quote}

To answer this question, we use the HydroDam environment, where a pumped hydro dam trades electricity by pumping water up or releasing water through a turbine. The environment dynamics, action set, reward definition, and DDQN architecture are fixed. Only the state representation given to the RL agent is varied during the ablation studies. We measure generalization by testing whether a policy selected on validation data remains effective on later prices from the same market and on unseen European electricity market zones. In ENTSO-E data, these market zones are called bidding zones: areas with one day-ahead price series, sometimes covering a whole country and sometimes a smaller price area.

Electricity markets can change a lot over time and between market zones. Price levels, volatility, negative prices, and daily price patterns can all differ from the data used for training. This makes feature design especially important. Absolute features directly describe the current price level and calendar time. Relative features compare the current price and calendar context with recent history, for example by measuring whether the current price is high or low compared with recent prices. Forecast features add an estimate of what prices may do next. These three feature types are useful for different reasons, but they may not generalize in the same way.

We study these feature choices with a controlled ablation. Policies are trained on the training period, selected using the validation period, and then evaluated on later same-market data and on other European market zones. This setup tests whether a feature set only works in the regime where it was selected, or whether it helps the policy keep working when the market changes.

The paper makes three contributions:
\begin{itemize}
  \item We formulate HydroDam as a controlled testbed for studying state representation in model-free deep RL for energy trading.
  \item We compare \texttt{absolute}, \texttt{relative}, and \texttt{forecast} state representations while keeping the environment and DDQN architecture fixed.
  \item We evaluate validation-selected policies on later same-market prices and unseen European market zones, showing that state representation strongly affects how well the policy keeps working when market conditions change.
\end{itemize}

A state with only absolute prices may learn thresholds that work in a calm historical period but fail when prices become negative or much more volatile. A state with only relative features can adapt better to changing price levels, but it may hide how large the actual trading opportunity is: a small price difference and a very large price difference can look similar after normalization. Forecast features can help the agent see likely future prices, which could help if the RL agent doesn't have learned internal forecasts. These feature-design choices have a direct effect on the final trading policy.

The evaluation is designed around these market changes. For each feature set and seed, checkpoints are selected based on validation-set performance. The selected policies are then evaluated on the same-market test set and on held-out market zones that are never used for selection. This exposes failures caused by changes in price levels, volatility, negative prices, and daily price structure.

The main findings are that \texttt{absolute} calendar/price features validate well but transfer poorly, \texttt{relative} features help most when combined with \texttt{absolute} features, and combining feature families is much more reliable than using any single family alone. The \texttt{absolute + relative + forecast} model gives the strongest held-out-zone results in the reported evaluation.

\section{Related work}

\paragraph{RL for energy storage and smart-grid control.}
Energy storage arbitrage has often been studied with optimization and valuation approaches, sometimes assuming perfect or forecasted price information \citep{sioshansi2009storage}. RL is another possible approach when future prices are uncertain, because the controller can learn from repeated simulated charging and discharging decisions. François-Lavet et al. use deep reinforcement learning to operate storage in a Belgian microgrid and compare inputs such as past time series, season information, and production forecasts \citep{francoislavet2016microgrids}. Wang and Zhang formulate storage arbitrage as an RL problem and show that reward and state design matter under real-time price uncertainty \citep{wang2018storage}. Xu et al. use deep RL for storage arbitrage with real electricity prices, where the policy maps available market information to charging and discharging decisions \citep{xu2019storage}. More broadly, RL has been surveyed as a tool for decision-making and control in power systems and sustainable energy applications \citep{glavic2017rlpower, lu2020rlenergy}. These papers motivate RL for storage control, but they usually do not isolate which parts of the market state make the learned policy transfer better. Our paper keeps the environment and DDQN algorithm fixed and studies that feature choice directly.

\paragraph{Electricity-price forecasts.}
Forecasting is a large part of electricity-market research because storage arbitrage depends on future price differences. Electricity prices are difficult to predict because they contain strong seasonality, sudden spikes, changing volatility, and market-specific effects. Reviews and benchmarks for day-ahead electricity price forecasting emphasize that methods should be tested on several markets and compared against strong simple baselines, not only on one short sample \citep{lago2021forecasting}. Recent work on battery energy arbitrage also shows that adding multi-horizon forecasts to a deep RL controller can improve reward, even when the forecasts are imperfect \citep{sage2024forecast}. In our paper, forecasts are not treated as a separate forecasting contribution. They are one state-feature family that is compared with absolute and relative features inside the same DDQN setup.

\paragraph{State representation in RL.}
The state representation determines what information the RL agent can use. In tabular RL, the choice of features directly controls the state-space size. In deep RL, the network can handle larger inputs, but the selected features still guide what the model can learn. State representation learning has been studied as a way to build compact and useful control features \citep{lesort2018srl}, while DQN-style methods \citep{mnih2015dqn} and Double DQN \citep{vanhasselt2016deep} show that neural value functions can learn from high-dimensional inputs. In applied energy trading, the state is often engineered by hand: current price, recent prices, calendar variables, storage level, and sometimes forecasts are chosen by the researcher \citep{wang2018storage,xu2019storage,sage2024forecast}. This makes feature design part of the model, not just a preprocessing detail.

\paragraph{Generalization under market shift.}
RL policies can overfit to the environments used during training and selection. Prior work on RL generalization argues that using separate training and test environments is important for measuring whether a policy really transfers \citep{cobbe2019generalization}. Surveys of zero-shot generalization make the same point for deployment settings where the test environment can differ from the training environment \citep{kirk2023survey}. Electricity markets create this problem naturally: the same policy may face later price regimes, negative prices, crisis-period spikes, or different market zones. This paper follows that idea in a storage-trading setting. Policies are trained on one period, selected based on validation-set performance, then evaluated on a later same-market test set and on unseen ENTSO-E market zones.

\section{Problem formulation and environment}

\paragraph{Decision process.}
The task is treated as a sequential decision problem. At each hour, the agent sees the current reservoir level and a set of features, then chooses whether to pump, release, or hold. It does not know the future price sequence, so its decision has to be based on the information available at that time. This makes the problem different from simply predicting whether the next price will go up or down. A profitable action also depends on the current reservoir level, the remaining storage capacity and the value of saving water for later hours. The same price can therefore imply different actions depending on the storage state and the recent market context. This provides the motivation for adopting a reinforcement learning approach. 

\paragraph{HydroDam task.}
The controlled environment models a pumped hydroelectric storage facility. The upper reservoir has a capacity of $100{,}000$ m$^3$ and starts half full. Water can be pumped upward or released through a turbine at up to $18{,}000$ m$^3$ per hour. Pumping has efficiency $0.8$ and generation has efficiency $0.9$. The height difference is fixed at $30$ m, so a volume change maps to energy through $mgh$ with water density $1000$ kg/m$^3$ and $g=9.81$ m/s$^2$.

At each hour $t$, the agent observes only information available up to that time and chooses one of three actions: pump at maximum rate, hold, or release at maximum rate. The reward is immediate trading revenue:
\begin{equation}
r_t =
\begin{cases}
- p_t \Delta E_t / 0.8, & \text{pump},\\
\phantom{-}0.9\,p_t \Delta E_t, & \text{release},\\
0, & \text{hold},
\end{cases}
\end{equation}
where $p_t$ is the hourly electricity price and $\Delta E_t$ is the energy associated with the upward or downward water movement in storage, which takes into account the minimum and maximum storage capacities. As $p_t$ simply follows the energy market, both positive and negative prices are possible. This means that during negative prices, pumping can be immediately profitable because the operator is paid to consume energy. 

The environment is simplified: it does not model natural water inflow, maintenance costs, market impact, or ancillary-service revenues, and it uses fixed pumping and generation efficiencies instead of a detailed turbine model. With the physical rules and action space fixed, differences in performance mainly reflect the state representation used during that experiment.

\paragraph{Optimal profit.}
For each price series, the best profit that could be achieved if the full future price sequence were known in advance is also calculated. This is not a realistic trading policy, since real controllers do not know future prices. Instead, it gives an upper bound on how much profit is available in that dataset. In this paper, the policy performance is reported as a percentage of this dataset-specific optimal profit. This makes results more comparable across markets. For example, a low-volatility zone and a high-volatility crisis period can have very different raw profit opportunities. Reporting percentages asks how much of the available opportunity each policy captures, rather than only how many euros it earns.

The optimal profit is computed with dynamic programming over the reservoir volumes that can actually be reached from the half-full initial state. The reservoir has a capacity of $100{,}000$ m$^3$ and the maximum hourly flow is $18{,}000$ m$^3$. With the three fixed actions and clipping at the empty and full bounds, this gives 17 possible volume states. For each hour and volume state, the backward pass evaluates pump, hold, and release. A forward pass then recovers the best cumulative profit from the initial state.

\section{State representation design}

All DDQN states include normalized reservoir volume. The ablations vary the market features. We group these features into three main types: absolute, relative, and forecast features.

\paragraph{Absolute features.}
Absolute state features include price normalized by the 99.9th percentile in the training set, log-transformed price, weekend indicator, and sine/cosine encodings of hour of day and day of week. These features can help the agent learn daily and weekly price patterns, but they can also make the policy depend too strongly on the price levels seen during training.

\paragraph{Relative features.}
Relative features describe the current market situation using a recent historical window. The basic relative price features use 24-hour and 168-hour z-scores and a rolling price score based on recent quantiles. These features follow the same idea as the strong price-score heuristic: buy when the current price is low compared to recent prices and sell when it is high.

Relative features are based on recent prices rather than on the full historical price distribution. This makes them more adaptable when the overall price level changes, because the agent can still see whether the current price is high or low compared with recent hours. The downside is that relative features can hide the size of the actual price difference. A small spread in a calm market may look similar to a much larger spread in a volatile market, even though the trading opportunity is very different.

The relative family also includes relative calendar features. These compare the current calendar situation with recent prices from similar situations. For example, instead of only telling the agent that it is hour 18, a relative hour feature indicates whether hour 18 has recently been relatively expensive or cheap compared with other hours. Similar features are used for day of week, weekend status, and month.

These features are meant to adapt when daily or weekly price patterns change. A fixed hour-of-day feature can learn that a certain hour was usually expensive during training, even if that pattern changes later. A relative calendar feature gives the agent a more recent view of how that calendar context behaves.

Finally, the relative family includes additional price indicators computed from current and past prices only. This adds price-range position, price-to-moving-average ratios, relative strength index, and Bollinger-band position. These features give the agent more information about recent trend, volatility, and where the current price sits within a recent price band.

\paragraph{Forecast features.}
Forecast features add information about expected prices over the next 24 hours. The forecast is produced by an ensemble of five LightGBM models \citep{ke2017lightgbm}. The models predict price differences recursively, so each step in the 24-hour forecast is generated from information available at the decision time and from earlier forecasted steps.

The forecaster uses lagged price differences, 24-hour and 168-hour rolling statistics, relative hour rank, EWMA, MACD, volatility-regime indicators, day-of-year encodings, calendar variables, and the recursive-step index. Thus, the forecast block does not give the RL agent the relative features directly, but its forecast is still trained from several relative inputs. This is important when interpreting the forecast result, because part of the relative information can already enter the policy through the forecasting model.

From the forecast window, the state also includes a future price z-score and an uncertainty estimate from the ensemble. These help the agent by summarizing the forecast window: the z-score gives the agent a direct comparison of the current price against the forecast, while the uncertainty estimate shows when the ensemble forecasts disagree. The forecasting ensemble is trained on the training set, with validation used for early stopping. The saved forecaster has 24-hour recursive MAE/RMSE of 5.58/26.86 EUR/MWh on validation and 15.22/32.35 on the test set.

Forecast features are evaluated both alone and in combination with absolute and relative features. This comparison tests whether forecasts are useful by themselves, and whether they still benefit from giving relative features directly to the RL agent.

\section{Experimental setup}

\paragraph{Data.}
The training and validation data cover Belgium/Belpex day-ahead prices from 2007--2011. This period is split in time, using 2007--2009 for training and 2010--2011 for validation. The validation split is used for model and checkpoint selection. The same-market test set then covers Belgian day-ahead prices from 2012--2025. This test set is later than the training and validation data and contains much larger price movements, including negative prices and the 2021--2022 energy crisis.

\begin{figure}[H]
  \centering
  \includegraphics[width=\linewidth]{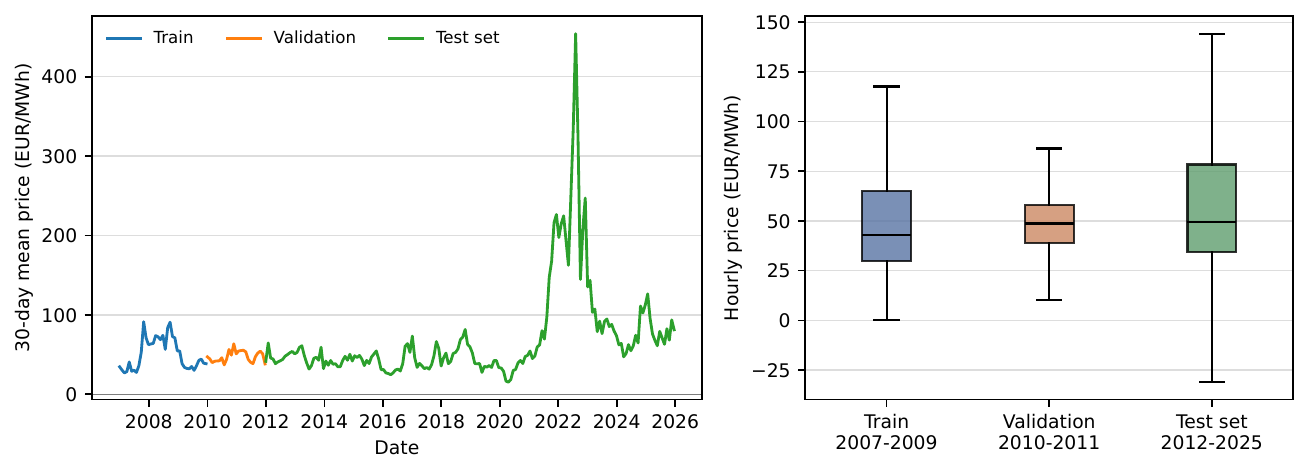}
  \caption{The 2012--2025 test set differs substantially from the 2007--2011 training/validation period, especially after the 2021--2022 energy crisis.}
  \label{fig:regime-shift}
\end{figure}

The first evaluation setting uses this test set to check whether the selected policies still work in a later period of the same market. The second evaluation setting tests whether the policies still work on other European markets, rather than only fitting the price levels of the training and validation data. For this, the same fixed policies are evaluated on 39 other European market zones, using each zone's longest consecutive available ENTSO-E range. The Belgian test set is built from public Belgian price data and ENTSO-E prices, and the cross-market panel comes from ENTSO-E bidding-zone files \citep{epfdata,entsoe}. Figure~\ref{fig:regime-shift} and Table~\ref{tab:dataset-characteristics} summarize the main differences between the training/validation period and the test set.

\begin{table}[H]
  \caption{Training, validation, and test-set price characteristics. Prices are EUR/MWh.}
  \label{tab:dataset-characteristics}
  \centering
  \begin{tabular}{llrrrrr}
    \toprule
    Dataset & Period & Hours & Mean & Std & Min & P95 \\
    \midrule
    Train & 2007--2009 & 26,304 & 50.60 & 40.09 & 0.01 & 104.68 \\
    Validation & 2010--2011 & 17,520 & 47.84 & 27.69 & 0.01 & 70.50 \\
    Test set & 2012--2025 & 122,736 & 70.44 & 72.46 & -500.00 & 206.02 \\
    \bottomrule
  \end{tabular}
\end{table}

\paragraph{Baselines and policies.}
The main non-RL baseline is a rolling price-score heuristic. This heuristic only uses past and current prices. At each hour, it looks at the previous 34 hours of prices and computes the 16th and 68th percentiles. The current price is then converted into a score,
\[
s_t = \frac{p_t - q_{16,t}}{q_{68,t} - q_{16,t}},
\]
where $p_t$ is the current price and $q_{16,t}$ and $q_{68,t}$ are the two rolling percentiles. A score of 0 means that the current price is at the recent low-price threshold, while a score of 1 means that it is at the recent high-price threshold. The heuristic pumps when $s_t \leq 0$, releases when $s_t \geq 1$, and otherwise holds. The window length and percentile thresholds were chosen using only the training data and then kept fixed for all evaluations. This gives a simple but strong baseline, because storage trading is largely about buying when the current price is low compared with recent prices and selling when it is high.

For each DDQN feature set and seed, the checkpoint is chosen using validation performance. The reported seed counts are listed in Appendix~\ref{app:reproducibility}. The test set and the held-out market zones are not used during checkpoint selection. After selection, the same fixed policies are evaluated on the 2012--2025 test set and on every zone in the 39-zone geographic panel, so the geographic results are not tuned separately for each market zone.

\paragraph{Evaluation implementation.}
For each dataset, the simulation starts with the reservoir half full and then steps through the price series hour by hour. At each hour, the DDQN policy chooses the action with the highest Q-value from the trained network. The DDQN policies, the heuristic, and the optimal-profit calculation all use the same dynamics and rewards. 

\paragraph{Agent.}
All learned policies use the same DDQN agent. The Q-network is a fully connected network with hidden layers of size $1024$-$1024$-$512$ and LeakyReLU activations. Training uses Huber loss, Adam with learning rate $10^{-4}$, discount factor $\gamma=0.99$, a replay buffer, target-network updates every 2500 steps, batch size 2048, and epsilon-greedy exploration. The full exploration schedule is given in Appendix~\ref{app:reproducibility}. These settings are kept fixed across feature sets, so differences between DDQN policies mainly come from the state representation.

\section{Results}

Table~\ref{tab:selection} first shows the same-market test setting. The \texttt{absolute} model has the strongest single-family validation result, reaching 78.9\% of optimal profit, but drops to 28.8\% on the 2012--2025 test set. This matches the concern from the feature design: absolute price and calendar features can fit the training/validation price scale too closely. The \texttt{relative} model is weaker by itself, reaching only 10.3\% on the test set. This suggests that relative features alone remove too much information about the size of the trading opportunity. The \texttt{forecast} model does better at 33.4\%, but it is still below the rolling price-score heuristic.

The main improvement comes from combining feature families. \texttt{absolute + relative} reaches 49.9\% on the test set, and \texttt{absolute + forecast} reaches a very similar 49.5\%, which is reasonable because the forecast is trained on relative price features as well. The full \texttt{absolute + relative + forecast} model is strongest at 55.6\%. This means that the \texttt{relative} and \texttt{forecast} families both help, but they help most when combined with absolute price and calendar information.

\begin{table}[H]
  \caption{Training, validation, and test-set performance. Values are reward as a percentage of optimal profit, reported as mean $\pm$ standard deviation over five seeds for DDQN policies.}
  \label{tab:selection}
  \centering
  \footnotesize
  \begin{tabular*}{0.98\linewidth}{@{\extracolsep{\fill}}lrrr}
    \toprule
    Representation & Train & Validation & Test set \\
    \midrule
    Price-score heuristic & 50.9 & 49.7 & 37.4 \\
    \texttt{absolute} & $75.3{\pm}1.5$ & $78.9{\pm}0.5$ & $28.8{\pm}4.6$ \\
    \texttt{relative} & $41.8{\pm}4.4$ & $26.7{\pm}2.7$ & $10.3{\pm}3.2$ \\
    \texttt{forecast} & $65.2{\pm}2.9$ & $54.9{\pm}2.2$ & $33.4{\pm}2.8$ \\
    \texttt{absolute + relative} & $86.8{\pm}0.9$ & $81.5{\pm}0.6$ & $49.9{\pm}1.6$ \\
    \texttt{absolute + forecast} & $88.1{\pm}0.7$ & $84.5{\pm}0.6$ & $49.5{\pm}3.0$ \\
    \texttt{relative + forecast} & $70.5{\pm}1.0$ & $63.5{\pm}0.9$ & $42.1{\pm}2.7$ \\
    \texttt{absolute + relative + forecast} & $89.4{\pm}0.8$ & $85.3{\pm}0.4$ & $55.6{\pm}1.3$ \\
    \bottomrule
  \end{tabular*}
\end{table}

\begin{figure}[H]
  \centering
  \includegraphics[width=\linewidth]{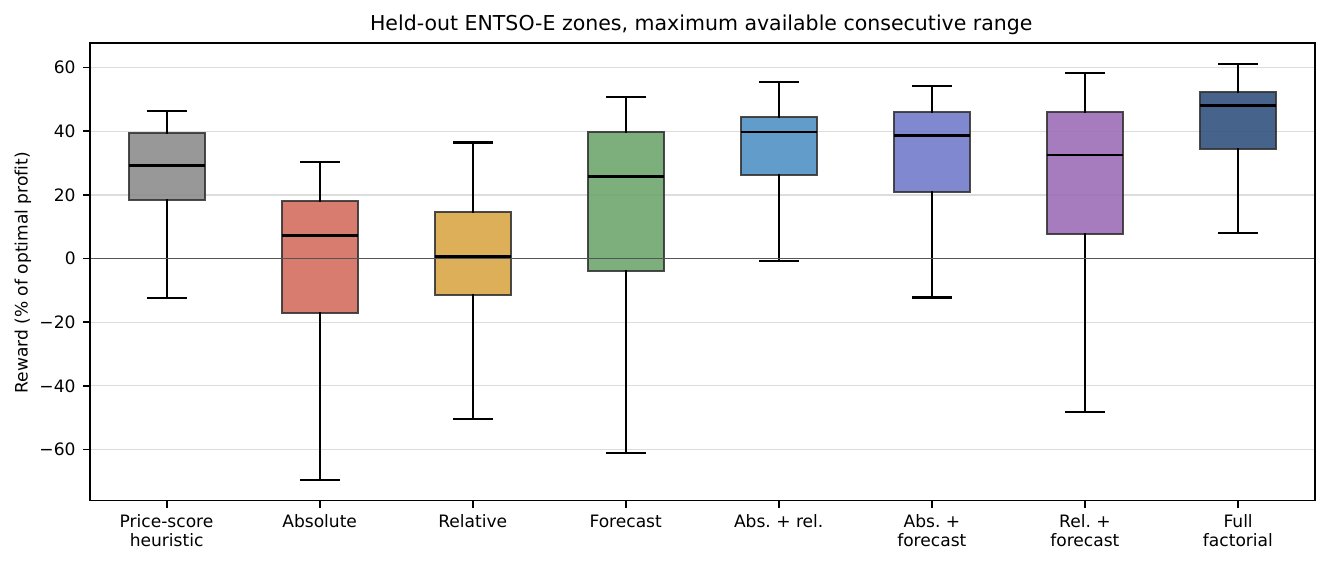}
  \caption{Cross-zone test on 39 ENTSO-E zones using the longest consecutive available range for each zone. Policies are fixed before evaluating these zones; DDQN boxes pool five seeds across zones.}
  \label{fig:cross-zone}
\end{figure}

The cross-zone results show the same pattern more strongly. Figure~\ref{fig:cross-zone} gives the full distribution across the 39 held-out zones. The rolling price-score heuristic reaches a median of 29.2\%, while \texttt{absolute-only} and \texttt{relative-only} DDQN policies are far below this, with medians of 5.7\% and 1.3\%. The \texttt{forecast} policy is the strongest single-family model, reaching 26.8\%, but it still does not beat the heuristic in the cross-zone median. As noted earlier, the forecast block is not completely independent of relative inputs, because the forecaster itself uses relative price information during training. Even so, giving only forecast features to the RL agent is not enough for strong transfer.

The two-feature combinations are the first DDQN policies that clearly pass the heuristic. \texttt{absolute + relative} reaches a cross-zone median of 39.8\%, while \texttt{absolute + forecast} reaches 40.0\%. \texttt{relative + forecast} is lower at 33.3\%, which shows that absolute price and calendar information still matters in most zones. The full \texttt{absolute + relative + forecast} model reaches the strongest aggregate result, with a median of 47.5\%, a mean of 40.2\%, and wins over the heuristic in 38.2 of 39 zones on average across seeds. Table~\ref{tab:cross-zone-summary} gives the exact aggregate values.

\begin{table}[H]
  \caption{Cross-zone test summary. Cross-zone values use each zone's longest consecutive available ENTSO-E range. Values are mean $\pm$ standard deviation over five seeds for DDQN policies; wins are out of 39 zones.}
  \label{tab:cross-zone-summary}
  \centering
  \footnotesize
  \begin{tabular*}{0.98\linewidth}{@{\extracolsep{\fill}}lrrr}
    \toprule
    Policy & Test set & Cross-zone median & Wins vs. heuristic \\
    \midrule
    Price-score heuristic & 37.4 & 29.2 & -- \\
    \texttt{absolute} & $28.8{\pm}4.6$ & $5.7{\pm}7.8$ & $1.0{\pm}0.0$ \\
    \texttt{relative} & $10.3{\pm}3.2$ & $1.3{\pm}3.1$ & $0.8{\pm}0.8$ \\
    \texttt{forecast} & $33.4{\pm}2.8$ & $26.8{\pm}3.0$ & $14.0{\pm}5.8$ \\
    \texttt{absolute + relative} & $49.9{\pm}1.6$ & $39.8{\pm}0.7$ & $36.0{\pm}1.6$ \\
    \texttt{absolute + forecast} & $49.5{\pm}3.0$ & $40.0{\pm}4.2$ & $32.4{\pm}5.0$ \\
    \texttt{relative + forecast} & $42.1{\pm}2.7$ & $33.3{\pm}2.4$ & $23.8{\pm}2.6$ \\
    \texttt{absolute + relative + forecast} & $55.6{\pm}1.3$ & $47.5{\pm}2.3$ & $38.2{\pm}0.4$ \\
    \bottomrule
  \end{tabular*}
\end{table}

\begin{figure}[H]
  \centering
  \makebox[\linewidth][c]{\includegraphics[width=1.08\linewidth]{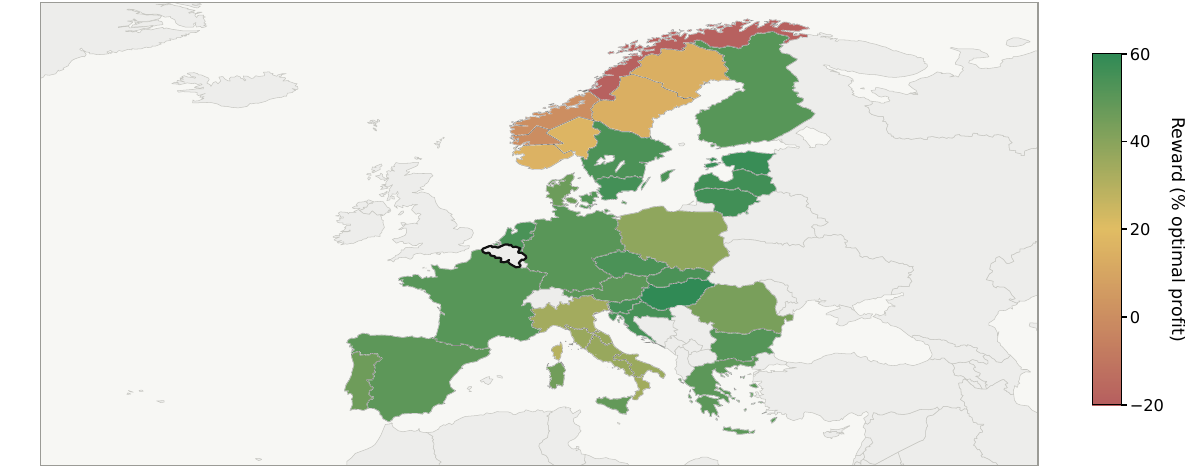}}
  \caption{Bidding-zone view of the \texttt{absolute + relative + forecast} cross-zone result. Colors show median reward as a percentage of optimal profit over five seeds. Belgium is outlined in black.}
  \label{fig:cross-zone-map}
\end{figure}

The per-zone results show that the feature families do not help in exactly the same places. Figure~\ref{fig:cross-zone-map} maps the \texttt{absolute + relative + forecast} result across the held-out bidding zones. The map shows several strong results near Belgium and in central Europe, which is plausible because neighbouring electricity markets are linked through cross-border exchange and can have more similar prices. However, the pattern is not only geographic: some farther zones also work well, while the Nordic zones remain weak. This model is the best DDQN policy in 34 of the 39 zones. Fourteen zones remain below 40\% of optimal profit, and seven of those are below 20\%. This shows that the combined state improves performance overall, but cross-zone results still vary strongly across market areas. The full per-zone panel is given in Appendix Table~\ref{tab:all-zones}, and Appendix Figure~\ref{fig:cross-zone-delta} shows representative examples.

There are also useful exceptions. In Romania, \texttt{relative + forecast} is strongest, reaching 50.2\%, while \texttt{absolute + relative + forecast} reaches 43.3\% (Appendix Table~\ref{tab:all-zones}). This is the only zone where the heuristic beats the full model. The hardest group is the Nordic zones. In NO1--NO5, SE1, and SE2, all policies remain far below the results in most other zones. These zones have smaller within-day price spreads, which makes them a harder setting for this short-duration arbitrage task. Together, these cases show that the combined state representation transfers much better overall, but it does not solve every market setting.

Overall, the cross-zone evaluation supports the same conclusion as the same-market test set, but more clearly. Single feature families are not reliable enough. Two-feature combinations are much stronger, especially when they include the \texttt{absolute} family. The best and most stable result comes from combining \texttt{absolute}, \texttt{relative}, and \texttt{forecast}.

\section{Discussion}

The results answer the research question directly: changing only the state representation changes the conclusion one would draw about DDQN. With absolute features alone, validation suggests a strong policy, but the held-out tests show poor transfer. With combined features, the same algorithm becomes much more competitive. This means the feature set is not a small preprocessing detail in this task, but a central part of the controller.

The \texttt{forecast} result has one important interpretation point. The forecast block is not fully separate from relative information, because the forecaster itself is trained using relative price inputs. This helps explain why \texttt{absolute + forecast} performs close to \texttt{absolute + relative}. At the same time, the best policy still uses direct relative features as well as forecast features. This suggests that the forecast gives useful forward-looking information, but it does not replace the need to show the DDQN the current relative price context directly.

The rolling price-score heuristic is an important part of this interpretation. It is simple, causal, and strong because storage arbitrage naturally depends on buying when the current price is low compared with recent prices and selling when it is high. A DDQN policy that cannot beat this baseline is not very useful in practice. The weak single-family DDQN results show that the neural network does not automatically recover a strong trading rule when the state is missing important information. The stronger combined-feature results show that DDQN becomes competitive when the state gives it the right market context.

The per-zone results also show that the \texttt{absolute + relative + forecast} state is not a complete solution to every market shift. Romania is an example where adding all feature families is not best, and the Nordic zones remain difficult for all policies. The main conclusion is therefore not that one state representation solves storage trading everywhere. It is that state representation has a large effect, and that combining scale, relative context, and forecast information transfers much better than relying on only one feature family.

If the goal were only to maximize deployment performance, a natural next step would be to train on more recent years and tune or fine-tune the controller for each target market. This paper does not do that because the goal is to compare state representations under the same fixed training and validation protocol.

\paragraph{Limitations.}
The study uses one storage environment and one value-based RL architecture, so the results should be read as evidence for this controlled DDQN setting rather than as a claim about all RL methods. Each DDQN feature set is rerun with five seeds, but all policies are still selected on the same validation period. The cross-zone test uses the longest consecutive available range for each zone, so the zones are not all evaluated on the same calendar years. The forecast model is also part of the controller, so forecast gains should be read as gains from the full forecast-plus-RL system. Within these limits, the evaluation shows that market-state features can create large differences on both later same-market prices and other European markets.

\section{Conclusion}

This paper studied how \texttt{absolute}, \texttt{relative}, and \texttt{forecast} features affect a DDQN controller for hydroelectric storage trading. The main result is that the state representation strongly changes both test-set performance and cross-zone transfer. Single feature families are not reliable enough: \texttt{absolute} features validate well but transfer poorly, while \texttt{relative-only} and \texttt{forecast-only} states stay below the price-score heuristic in the cross-zone median. Combining feature families is much stronger, and the \texttt{absolute + relative + forecast} state gives the best overall result. For model-free RL in energy storage trading, the state representation should therefore be treated as a central design choice: the agent needs price scale, recent relative price position, and likely short-term future prices to keep working when market conditions change.

\newpage

\bibliographystyle{unsrtnat}
\bibliography{references}

\appendix

\section{Reproducibility details}
\label{app:reproducibility}

Table~\ref{tab:checkpoint-protocol} lists the checkpoint-selection protocol for the reported policies. The action order in the Q-network is release, hold, pump. Rewards are divided by 1000 before replay-buffer insertion. Training uses 2048 parallel environments, batch size 2048, target-network updates every 2500 vector steps, and a five-step warm-up. Epsilon-greedy exploration starts at 1.0. After each training episode, epsilon is multiplied by 0.95 until it reaches a minimum value of 0.05. Each DDQN feature set is trained for 20 episodes with five seeds, and the checkpoint with the highest validation optimal-profit percentage is selected within each seed.

\begin{table}[htbp]
  \caption{Checkpoint-selection details for the reported DDQN policies.}
  \label{tab:checkpoint-protocol}
  \centering
  \resizebox{0.98\linewidth}{!}{%
  \begin{tabular}{lrrl}
    \toprule
    Representation & State dim. & Seeds & Selection metric \\
    \midrule
    \texttt{absolute} & 8 & 5 & Validation optimal-profit percentage \\
    \texttt{relative} & 14 & 5 & Validation optimal-profit percentage \\
    \texttt{forecast} & 28 & 5 & Validation optimal-profit percentage \\
    \texttt{absolute + relative} & 21 & 5 & Validation optimal-profit percentage \\
    \texttt{absolute + forecast} & 35 & 5 & Validation optimal-profit percentage \\
    \texttt{relative + forecast} & 41 & 5 & Validation optimal-profit percentage \\
    \texttt{absolute + relative + forecast} & 48 & 5 & Validation optimal-profit percentage \\
    \bottomrule
  \end{tabular}}
\end{table}

The absolute block uses reservoir volume, normalized/log price, weekend, and hour and day-of-week sine/cosine features. The relative block uses recent price and calendar context: relative hour, relative day of week, relative weekend, relative month, 24-hour and 168-hour price z-scores, a rolling price score, price-range position, price-to-moving-average ratios, RSI, and Bollinger-band position. The forecast block uses the recursive 24-hour forecast, future z-score, and forecast uncertainty. The reported DDQN policies cover all non-empty combinations of these three blocks.

\section{Held-out-zone panel}

Table~\ref{tab:all-zones} reports the full market-zone panel used for the cross-zone summary. The policy columns are reward as percentage of each zone's optimal profit. Zone names are ENTSO-E bidding-zone identifiers; some Italian and Nordic zones are sub-national bidding zones.

\begin{table}[H]
  \caption{ENTSO-E zones used in the maximum-available cross-zone test. Days gives the length of the evaluated consecutive range. Policy values are medians over five seeds.}
  \label{tab:all-zones}
  \centering
  \resizebox{0.98\linewidth}{!}{%
  \begin{tabular}{lrrrrrrrrr}
    \toprule
    Zone & Days & Heuristic & \texttt{Abs.} & \texttt{Rel.} & \texttt{Forecast} & \texttt{Abs. + rel.} & \texttt{Abs. + forecast} & \texttt{Rel. + forecast} & \texttt{Full} \\
    \midrule
    AT & 4018 & 30.3 & 0.6 & 1.1 & 28.8 & 39.1 & 43.0 & 35.6 & 49.7 \\
    BG & 3348 & 46.4 & 14.3 & 5.4 & 45.1 & 49.1 & 40.7 & 49.3 & 51.3 \\
    BSP & 4018 & 39.3 & 14.0 & 16.1 & 37.3 & 46.6 & 46.4 & 46.0 & 55.2 \\
    CALA & 1826 & 21.5 & -17.0 & -6.5 & -17.5 & 24.1 & 16.4 & 8.2 & 34.1 \\
    CNOR & 4014 & 17.2 & -23.8 & -6.1 & -3.3 & 26.7 & 21.6 & 7.1 & 36.2 \\
    COAC & 4014 & 33.3 & -1.2 & -2.0 & 8.0 & 40.2 & 29.3 & 24.4 & 44.8 \\
    CORS & 4014 & -0.5 & 17.2 & -0.6 & -49.5 & 22.6 & 26.0 & -9.8 & 29.6 \\
    CSUD & 4014 & 19.8 & -25.4 & -5.3 & -3.4 & 27.9 & 22.7 & 10.7 & 36.4 \\
    CZ & 4018 & 36.5 & 11.8 & 13.6 & 34.8 & 43.4 & 44.2 & 44.0 & 53.5 \\
    DE\_smart\_mixed & 4014 & 30.6 & 12.0 & -1.6 & 33.7 & 40.3 & 44.1 & 37.4 & 50.4 \\
    DK1 & 4018 & 26.2 & 9.8 & 5.0 & 29.8 & 34.9 & 39.4 & 35.3 & 46.0 \\
    DK2 & 4018 & 29.2 & 16.1 & 11.5 & 35.8 & 38.9 & 45.2 & 42.0 & 51.4 \\
    EE & 4018 & 41.9 & 20.6 & 31.9 & 44.3 & 48.1 & 52.4 & 53.4 & 57.6 \\
    ES & 4018 & 22.2 & 5.8 & -12.1 & 15.7 & 40.4 & 45.2 & 21.0 & 49.5 \\
    FI & 4018 & 28.8 & 19.6 & 23.7 & 44.0 & 40.1 & 50.4 & 45.6 & 50.9 \\
    FR & 4014 & 30.4 & 19.6 & 1.2 & 31.3 & 42.7 & 43.8 & 32.5 & 50.8 \\
    GREE & 4018 & 41.0 & -6.8 & 9.5 & 22.4 & 41.6 & 35.3 & 37.6 & 49.8 \\
    HR & 2997 & 40.6 & 13.2 & 3.4 & 39.7 & 51.6 & 45.4 & 46.4 & 54.4 \\
    HU & 4018 & 44.5 & 15.6 & 25.3 & 48.2 & 50.2 & 49.5 & 54.2 & 59.4 \\
    LT & 4018 & 41.6 & 18.4 & 27.7 & 42.0 & 46.5 & 49.7 & 50.7 & 55.9 \\
    LV & 4018 & 41.9 & 18.1 & 28.4 & 41.7 & 46.6 & 49.6 & 50.6 & 55.9 \\
    NL & 4014 & 40.1 & 20.3 & 0.6 & 31.9 & 49.9 & 46.2 & 42.2 & 53.9 \\
    NO1 & 4018 & -13.8 & -36.8 & -53.3 & -23.6 & 7.8 & 8.7 & -20.9 & 16.3 \\
    NO2 & 4018 & -12.0 & -38.7 & -55.4 & -29.4 & 6.4 & 8.2 & -23.8 & 15.1 \\
    NO3 & 4018 & -12.3 & -48.8 & -39.0 & -12.1 & -9.9 & 3.5 & -15.5 & -0.0 \\
    NO4 & 4018 & -28.9 & -80.3 & -63.5 & -35.2 & -27.0 & -12.3 & -38.5 & -18.9 \\
    NO5 & 4018 & -30.8 & -62.3 & -81.7 & -50.7 & -10.4 & -9.7 & -48.1 & -0.4 \\
    NORD & 4014 & 13.2 & -26.2 & -9.3 & -11.1 & 22.4 & 18.7 & -1.3 & 33.9 \\
    PL & 4014 & 29.1 & -4.7 & 4.4 & 28.3 & 29.6 & 27.7 & 31.9 & 38.2 \\
    PT & 4018 & 20.6 & 2.1 & -15.6 & 10.8 & 38.1 & 41.7 & 16.3 & 45.9 \\
    RO & 4018 & 45.1 & 8.2 & 25.4 & 43.7 & 40.5 & 32.1 & 50.2 & 43.3 \\
    SARD & 4014 & 33.3 & -1.3 & -2.6 & 7.8 & 40.4 & 29.5 & 24.3 & 45.0 \\
    SE1 & 4018 & -1.6 & -23.3 & -17.9 & 7.6 & 1.3 & 17.9 & 2.9 & 14.0 \\
    SE2 & 4018 & -0.6 & -21.6 & -15.0 & 7.2 & 2.0 & 18.7 & 3.5 & 13.8 \\
    SE3 & 4018 & 26.6 & 21.3 & 18.6 & 46.9 & 42.7 & 52.2 & 48.0 & 53.4 \\
    SE4 & 4018 & 31.3 & 23.4 & 19.6 & 45.5 & 45.2 & 53.3 & 49.6 & 55.3 \\
    SICI & 4014 & 29.7 & -6.3 & 2.3 & 13.7 & 42.2 & 36.8 & 30.5 & 48.3 \\
    SK & 4018 & 39.8 & 13.9 & 19.6 & 39.8 & 45.6 & 44.1 & 47.9 & 53.5 \\
    SUD & 4014 & 21.7 & -25.0 & -5.3 & -6.0 & 26.3 & 22.2 & 11.9 & 35.8 \\
    \bottomrule
  \end{tabular}}
\end{table}

\begin{figure}[H]
  \centering
  \makebox[\linewidth][c]{\includegraphics[width=1.12\linewidth]{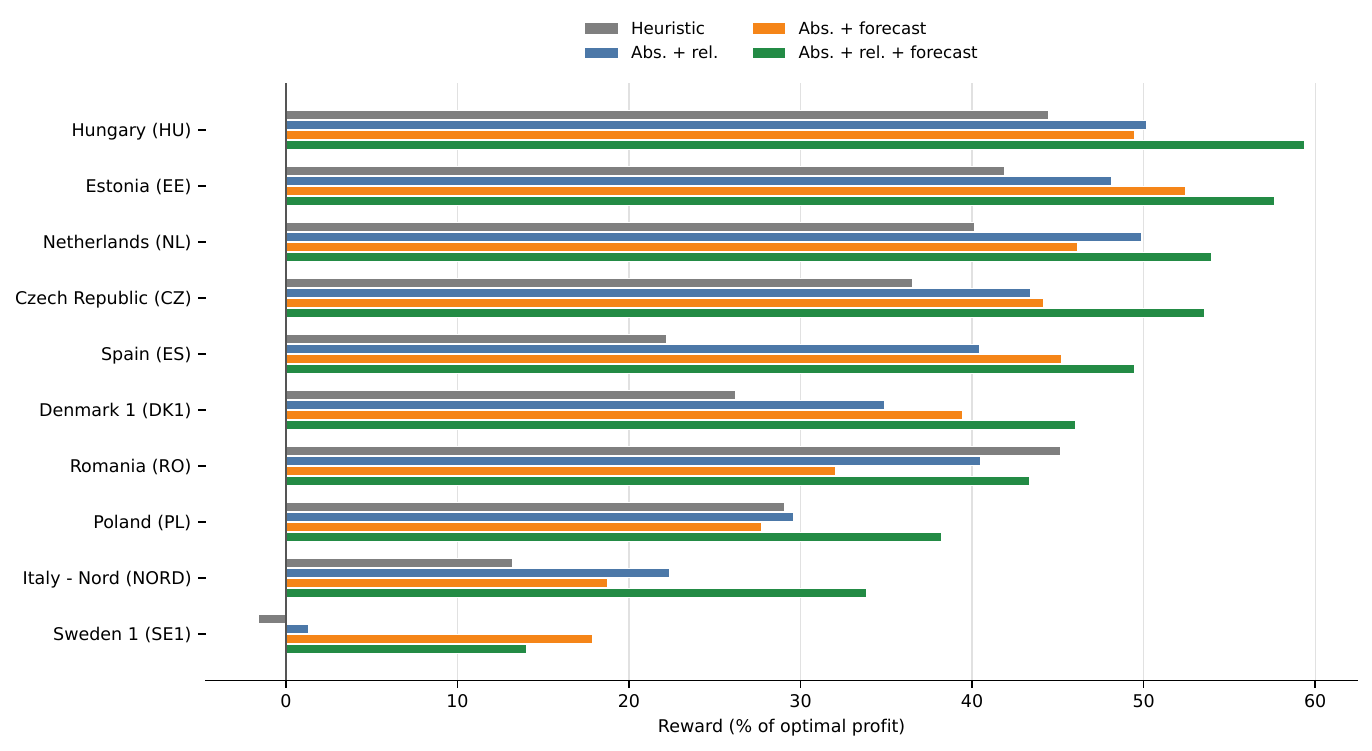}}
  \caption{Representative cross-zone comparison for market zones chosen to cover lower, middle, strong, and exception cases from the full 39-zone panel. Bars show median reward as a percentage of optimal profit over five seeds for the price-score heuristic and the main combined DDQN feature sets.}
  \label{fig:cross-zone-delta}
\end{figure}

\end{document}